%% file: main.tex
\documentclass{article}

\usepackage{microtype}
\usepackage{graphicx}
\usepackage{booktabs}
\usepackage{longtable}
\usepackage{tabularx}
\usepackage{adjustbox}
\usepackage[utf8]{inputenc}
\usepackage[T1]{fontenc}
\usepackage{amsmath,amssymb,amsfonts}
\usepackage{hyperref}
\usepackage{dblfloatfix}
\usepackage{xcolor}
\usepackage{enumitem}

\usepackage[preprint]{icml2026}

\setcitestyle{numbers,square,comma}

\icmltitlerunning{Story2Proposal:  A Scaffold for Structured Scientific Paper Writing}

\begin{document}

\twocolumn[
\icmltitle{Story2Proposal: A Scaffold for Structured Scientific Paper Writing}

\icmlsetsymbol{equal}{*}

\begin{icmlauthorlist}
\icmlauthor{Zhuoyang Qian${}^*$}{anon}
\icmlauthor{Wei Shi${}^*$}{anon}
\icmlauthor{Xu Lin${}^*$}{anon}
\icmlauthor{Li Ling${}^*$}{anon}
\icmlauthor{Meng Luo${}^*$}{anon}

\icmlauthor{Ziming Wang}{anon}
\icmlauthor{Tengyue Xu}{anon}
\icmlauthor{Gaoge Liu}{anon}
\icmlauthor{Zhentao Zhang}{anon}
\icmlauthor{Shuo Zhang}{anon}
\icmlauthor{Ziqi Wang}{anon}
\icmlauthor{Zheng Feng}{anon}
\icmlauthor{Yan Luo}{anon}
\icmlauthor{Shu Xu}{anon}
\icmlauthor{Yongjin Chen}{anon}
\icmlauthor{Zhiwei Zhang}{anon}
\icmlauthor{Zhibo Feng}{anon}
\icmlauthor{Zhuo Chen}{anon}
\icmlauthor{Bruce Yuan}{anon}
\icmlauthor{Biao Wu${}^\dagger$}{anon}
\icmlauthor{Harry Wang${}^{\ddagger}$}{anon}
\icmlauthor{Kris Chen${}^{\ddagger}$}{anon}
\end{icmlauthorlist}

\icmlaffiliation{anon}{AgentAlpha Team}
\icmlcorrespondingauthor{Harry Wang}{wanghuacan17@mails.ucas.ac.cn}
\icmlcorrespondingauthor{Kris Chen}{chenronghao@alumni.pku.edu.cn}

\icmlkeywords{Scientific Writing, Multi-Agent Systems, Paper Writing Scaffold}

\vspace{0.5em}
\begin{center}
{\small
${}^*$ Equal contribution \qquad
${}^\dagger$ Project lead \qquad
${}^\ddagger$ Corresponding authors
}
\end{center}

\vspace{0.5em}
\begin{center}
{\color{gray}\emph{This manuscript, including its writing and figures, was primarily generated by the Story2Proposal Agent. Human contributors focused on designing and implementing the agent system, as well as collecting and organizing experimental results, without direct involvement in the writing process. As such, any imperfections in writing reflect the current capabilities of the system rather than extensive manual polishing.}}
\end{center}

\vskip 0.3in
]

\printAffiliationsAndNotice{}

\begin{abstract}
Generating scientific manuscripts requires maintaining alignment between narrative reasoning, experimental evidence, and visual artifacts across the document lifecycle. Existing language-model generation pipelines rely on unconstrained text synthesis with validation applied only after generation, often producing structural drift, missing figures or tables, and cross-section inconsistencies. We introduce Story2Proposal, a contract-governed multi-agent framework that converts a research story into a structured manuscript through coordinated agents operating under a persistent shared visual contract. The system organizes architect, writer, refiner, and renderer agents around a contract state that tracks section structure and registered visual elements, while evaluation agents supply feedback in a generate–evaluate–adapt loop that updates the contract during generation. Experiments on tasks derived from the Jericho research corpus show that Story2Proposal achieved an expert evaluation score of 6.145 versus 3.963 for DirectChat (+2.182) across GPT, Claude, Gemini, and Qwen backbones. Compared with the structured generation baseline Fars, Story2Proposal obtained an average score of 5.705 versus 5.197, indicating improved structural consistency and visual alignment.
\end{abstract}

\input{sections/introduction}
\input{sections/related_work}

\input{sections/method}
\input{sections/experiments}
\input{sections/analysis}
\input{sections/conclusion}

\bibliographystyle{unsrtnat}
\bibliography{paper}


\newpage




\end{document}

%% file: sections/introduction.tex
\section{Introduction}

Academic paper generation represents a fundamental challenge in long-form text synthesis \cite{lu2024ai, gridach2025agentic, chen2025ai4research, eger2025transforming, zheng2025automation}, where maintaining global coherence, argumentative consistency, and structural integrity across thousands of words remains difficult for current language models. Consider a researcher attempting to generate a complete conference paper from a high-level research concept: the introduction must establish motivation and contributions, the method section must provide technical details that align with those contributions, the experiments must validate the exact claims stated earlier, and the related work must position the approach against cited baselines without contradicting the methodology \cite{gottweis2025towards, weng2025deepscientist, luo2025llm4sr, zhu2026autofigure,xu2026idea2paper}. This requirement for cross-section consistency, claim-evidence alignment, and citation grounding distinguishes academic writing from other long-form generation tasks. Modular approaches to complex reasoning \cite{khot2022decomposed, boiko2023autonomous, m2024augmenting} and controllable generation frameworks \cite{zhang2022automatic, weng2024cycleresearcher, tang2025ai} have demonstrated the value of intermediate representations for maintaining coherence, yet existing systems fail to provide the semantic control mechanisms necessary for paper-length document generation.

Current approaches to automated academic writing suffer from fundamental limitations in maintaining consistency across paper-length documents. Direct prompting methods \cite{wang2024autosurvey, shao2024assisting, liang2025surveyx, susnjak2025automating, koncel2019text} generate papers end-to-end without intermediate control mechanisms, leading to section drift where later sections diverge from the initial problem framing, claim-experiment misalignment where stated contributions lack corresponding empirical validation, citation-argument disconnection where references fail to ground specific technical claims, and inter-section contradictions where methodology descriptions conflict with experimental setups. Single-agent long-form generation systems \cite{agashe2024agent, yang2023doc, bai2024longwriter, huot2024agents, wan2025cognitive} lack the specialized expertise needed for different paper sections, producing generic content that fails to capture domain-specific argumentation patterns. Outline-based approaches \cite{yang2022re3, kim2024navigating, liang2024integrating, wang2024dome, xiong2025beyond} provide high-level structure but offer insufficient semantic grounding to maintain argumentative coherence across sections. Agent frameworks with planning capabilities \cite{, hong2023metagpt, wu2023autogen, schmidgall2025agent} demonstrate coordination mechanisms but do not address the provenance tracking required to ensure generated text remains faithful to source concepts. Controllable generation methods \cite{feng2022training, dathathri2020plug, yang2021fudge, mudgal2024controlled, liu2024improving} establish the need for intermediate control but do not provide the structured semantic layer necessary for academic writing, where each claim must trace back to explicit research contributions and each experiment must validate stated hypotheses.

To address these limitations, we propose Story2Proposal, a contract-governed multi-agent framework for automated scientific manuscript generation that integrates collaborative agents with persistent structural governance. Instead of treating manuscript production as unconstrained text synthesis, Story2Proposal models it as a contract-governed construction process in which structural and visual obligations are explicitly represented and continuously enforced through a persistent shared visual contract. The framework coordinates four specialized agents: an architect agent that transforms a research story into a structured blueprint and initializes the contract state, a writer agent that generates section drafts under contract constraints, a refiner agent that improves coherence and global narrative alignment, and a renderer agent that materializes figures, tables, and LaTeX structure consistent with the contract. Throughout generation, the contract state records section structure, registered visual artifacts, and validation rules. Evaluation agents analyze intermediate outputs for reasoning quality, data fidelity, and visual consistency, producing feedback that updates the contract state and guides subsequent generation steps within a generate--evaluate--adapt loop.

We evaluate the framework across multiple large language model settings and manuscript-generation scenarios derived from the Jericho research corpus. Comparative experiments against strong baselines show that contract-governed generation improves structural consistency, visual integration, and overall manuscript coherence. The results indicate that persistent structural contracts stabilize multi-stage generation and reduce structural drift during iterative rewriting. Qualitative analysis further shows that the persistent shared visual contract preserves figure placement and maintains alignment between narrative reasoning and supporting artifacts during the document lifecycle.

Our main contributions are:

\begin{itemize}[leftmargin=*, labelsep=0.5em]
\item We introduce Story2Proposal, a contract-governed multi-agent framework that coordinates architect, writer, refiner, and renderer agents through a persistent shared visual contract to maintain structural and visual consistency during scientific manuscript generation. Unlike DirectChat, which performs single-stage prompt-based generation without persistent constraints, Story2Proposal maintains a contract state that explicitly tracks section obligations and visual artifacts.

\item We design a generate--evaluate--adapt mechanism that integrates evaluation agents for reasoning validation, data fidelity checking, and visual coherence assessment. In contrast to FARS, which applies structured generation procedures without continuous verification feedback, the proposed framework dynamically updates the contract state using evaluation signals to guide subsequent generation.

\item We present an empirical evaluation with expert reviewers across multiple language model backbones and manuscript scenarios, demonstrating improved structural integrity, visual consistency, and overall manuscript robustness compared with DirectChat and FARS. These results suggest that contract-governed generation improves the reliability of automated scientific writing systems.
\end{itemize}

%% file: sections/related_work.tex
\section{Related Work}

\subsection{Modular Approaches to Complex Text Generation}

Recent work \cite{yamada2025ai, mitchener2025kosmos, ren2025towards} has explored decomposing complex generation tasks into modular subtasks to improve controllability and output quality. Decomposed Prompting breaks down complex tasks into simpler subproblems that can be solved independently and then composed, demonstrating improved performance on multi-step reasoning tasks through explicit task decomposition \cite{shang2024agentsquare, li2025agentswift, luo2025dr, xi2025survey, luo2026unveiling}. Take a Step Back proposes evoking reasoning via abstraction by first generating high-level principles before tackling specific instances, showing that intermediate abstraction layers can guide more coherent problem-solving. In the visual domain, Training-Free Structured Diffusion introduces compositional guidance mechanisms that maintain structural constraints during image generation through explicit intermediate representations \cite{khot2022decomposed, pang2025paper2poster, zheng2025pptagent}. While these approaches demonstrate the value of intermediate representations and modular decomposition, they operate within single-agent frameworks that lack specialized expertise for different subtask types. Furthermore, these methods do not maintain bidirectional provenance tracking between intermediate representations and final outputs, making it difficult to trace which parts of the generated content derive from specific intermediate decisions \cite{huang2025deep, hang2025beyond, shi2025deep}. However, these modular approaches do not address the challenge of maintaining global consistency across multiple interdependent sections in long-form documents, which our method resolves by introducing structured research stories as a semantic intermediate layer with explicit provenance tracking between story fields and generated paper sections.

\subsection{Multi-Agent Architectures for Structured Tasks}

\begin{figure*}[t]
  \centering
  \includegraphics[width=0.8\textwidth]{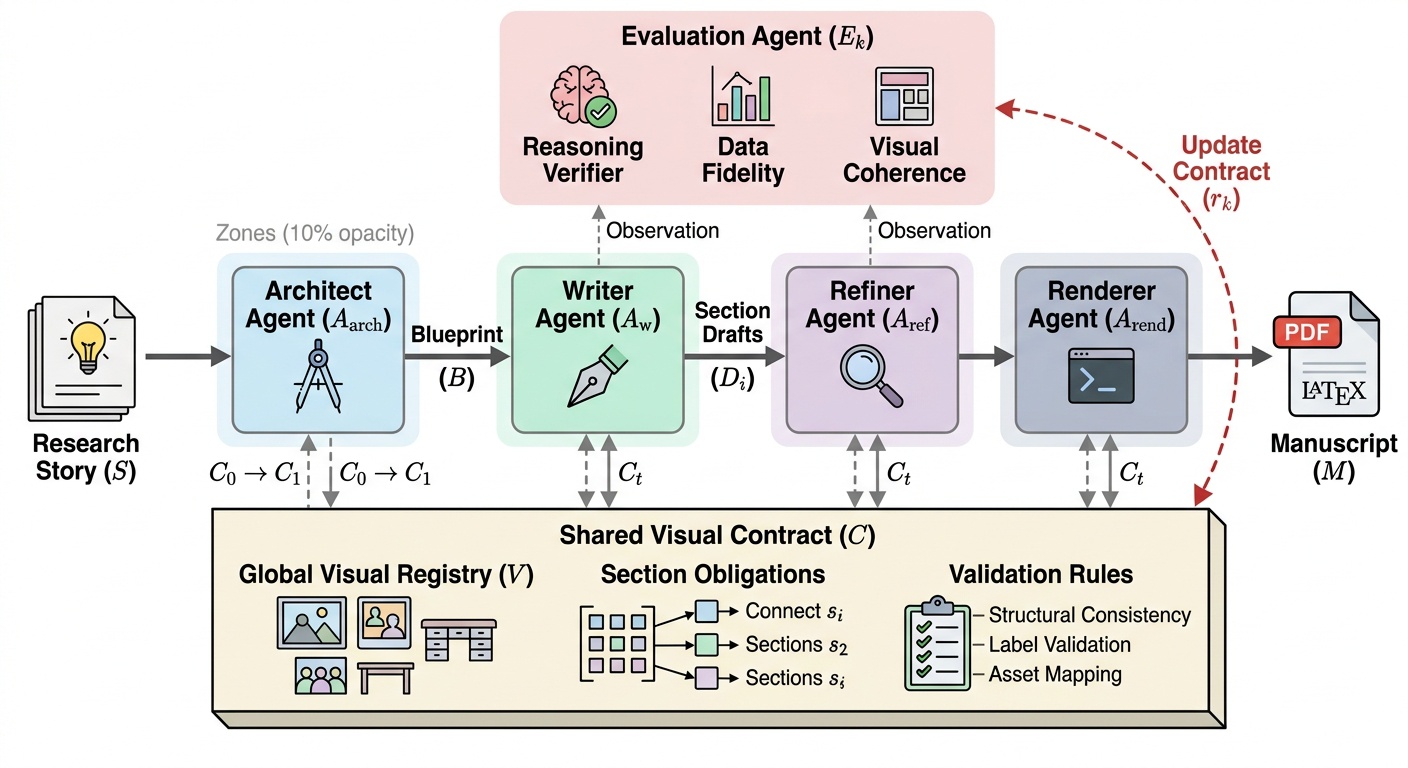}
  \caption{System overview of Story2Proposal. Given a research story as input, the framework coordinates architect, writer, refiner, and renderer agents through a persistent shared visual contract that records section structure, registered visual artifacts, and validation rules. Evaluation agents inspect intermediate outputs and feed corrective signals back to the contract state, enabling provenance-aware planning, structurally constrained drafting, global refinement, and deterministic rendering of the final manuscript.}
  \label{fig:system_overview}
\end{figure*}

Multi-agent systems \cite{tran2025multi, sun2025multi, jin2025comprehensive, hong2023metagpt, wu2023autogen} have emerged as a paradigm for tackling complex tasks through specialized agent collaboration. AgentSquare \cite{shang2024agentsquare} proposes automatic agent search in modular design spaces, demonstrating that composing specialized agents with distinct capabilities can outperform monolithic models on complex reasoning tasks \cite{zheng2023take, du2024improving}. Socratic Models introduces a framework for composing zero-shot multimodal reasoning by orchestrating multiple language models, each specialized for different modalities, through structured dialogue protocols \cite{zeng2022socratic}. Gamma Sampling provides fine-grained control over language model outputs without training by dynamically adjusting sampling distributions based on constraint satisfaction, enabling controllable generation through inference-time intervention \cite{feng2022training, liu2021dexperts}. Confronting Reward Model Overoptimization addresses quality control in language model outputs through constrained reinforcement learning, establishing mechanisms for maintaining output quality during iterative refinement \cite{moskovitz2023confronting, gao2023scaling}. These multi-agent and controllable generation approaches demonstrate the benefits of specialization and coordination, yet they lack grounding in structured intermediate semantic representations that explicitly encode domain-specific knowledge \cite{pan2024unifying}. Moreover, existing multi-agent systems for text generation do not maintain revision memory across iterative refinement cycles, leading to inconsistent modifications when addressing critique feedback \cite{madaan2023self, shinn2023reflexion}. However, these systems do not provide provenance-grounded coordination mechanisms that trace generated content back to specific semantic fields in structured input representations, which our approach addresses through discourse planning that maps story fields to section-level generation tasks and maintains cross-section consistency through shared claim-evidence maps.

%% file: sections/method.tex
\section{Method}

Story2Proposal models automated scientific manuscript generation as a coordinated multi-agent ecosystem governed by a shared structural contract, reflecting recent perspectives that frame large language models as cooperative agents capable of coordinating specialized functions to accomplish complex tasks \cite{zhang2024offlinetraining}. Rather than treating generation as a single-pass language modeling task, the framework decomposes manuscript creation into specialized agents that operate over a persistent representation of structural, visual, and citation obligations. \cite{hausknecht2020interactivefictiongamescolossal} Four generation agents—architect, writer, refiner, and renderer—interact with evaluation agents that assess reasoning quality, data fidelity, and visual consistency while generation proceeds. These agents communicate through an evolving contract state that constrains generation and adapts in response to evaluation feedback. The overall architecture is illustrated in Figure \ref{fig:system_overview}.

This design follows the paradigm of role-specialized collaborative language agents, which improve complex task decomposition and reliability compared with monolithic generation systems \cite{hong2023metagpt}. Multi-agent ecosystems enable structured coordination in which different agents focus on planning, generation, and verification stages of reasoning \cite{chen2024autoagents}. Story2Proposal extends this paradigm by introducing a persistent shared visual contract that enforces document-level structural and visual constraints throughout the generation pipeline.

\par\noindent
\begin{minipage}{\columnwidth}
\centering
\normalsize
\setlength{\tabcolsep}{3pt}
\begin{tabularx}{\columnwidth}{@{}l X@{}}
\toprule
Symbol & Meaning \\
\midrule
$S$ & input research story and supporting evidence \\
$C$ & shared visual contract encoding global and section-level visual obligations \\
$C_{t}$ & state of the visual contract at generation stage t \\
$A_{arch}$ & architect agent generating the manuscript blueprint \\
$A_{w}$ & writer agent producing section drafts under contract constraints \\
$A_{ref}$ & refiner agent performing global coherence alignment and compression \\
$A_{rend}$ & renderer agent producing final LaTeX with deterministic validation \\
$E_{k}$ & evaluation agent k assessing scientific reasoning, data fidelity, or visual coherence \\
$M$ & generated manuscript artifact \\
$V$ & set of required visual elements (figures and tables) \\
$s_{i}$ & manuscript section i \\
$r_{k}$ & evaluation feedback signal produced by evaluation agent k \\
$\Delta$ & performance delta between compared generation methods \\
\bottomrule
\end{tabularx}
\captionsetup{type=table}
\captionof{table}{Notation used throughout the method.}
\label{tab:notation}
\end{minipage}
\par

\subsection{Problem Formulation}

Let $S$ denote an input research story consisting of narrative descriptions, experimental evidence, and contextual information describing a scientific contribution. The objective is to transform $S$ into a structured manuscript artifact $M$ suitable for academic publication.

Unlike conventional generation pipelines that treat a manuscript as an unconstrained text sequence, Story2Proposal introduces an explicit contract governing structural and visual requirements. The contract specifies section organization, required visual artifacts, and reference relationships that must be satisfied during generation.

Let $C$ denote the shared visual contract and $V$ the set of required visual elements such as figures and tables. Let $\{s_i\}$ denote the ordered set of manuscript sections. The contract specifies how elements of $V$ must appear within sections $s_i$ and how they are referenced.

Manuscript construction proceeds through a sequence of contract-governed transformations:

\vspace{-3mm}
\begin{equation}
M = A_{rend}(A_{ref}(A_w(A_{arch}(S, C_0), C_1), C_2), C_3)
\end{equation}

where $A_{arch}$, $A_w$, $A_{ref}$, and $A_{rend}$ represent the architect agent, writer agent, refiner agent, and renderer agent respectively. The contract evolves across stages as $\{C_t\}$.

Evaluation agents provide feedback during generation. Given evaluation signals $\{r_k\}$ from evaluation agents, the contract state is updated as

\vspace{-3mm}
\begin{equation}
C_{t+1} = \text{Update}(C_t, \{r_k\})
\end{equation}

The final manuscript therefore emerges from coordinated agent interactions constrained by a continuously updated contract rather than a static prompt.

\subsection{Notation}

We define the key symbols used throughout the method in Table~\ref{tab:notation}. The notation describes the input research story, the shared visual contract and its state transitions, the generation agents, the evaluation agents, and the produced manuscript artifact.

\subsection{Shared Visual Contract}

The shared visual contract $C$ provides the structural mechanism that coordinates generation across agents. It acts as a persistent representation that records all visual and structural obligations associated with the manuscript.

\begin{figure*}[t]
  \centering
  \includegraphics[width=0.8\textwidth]{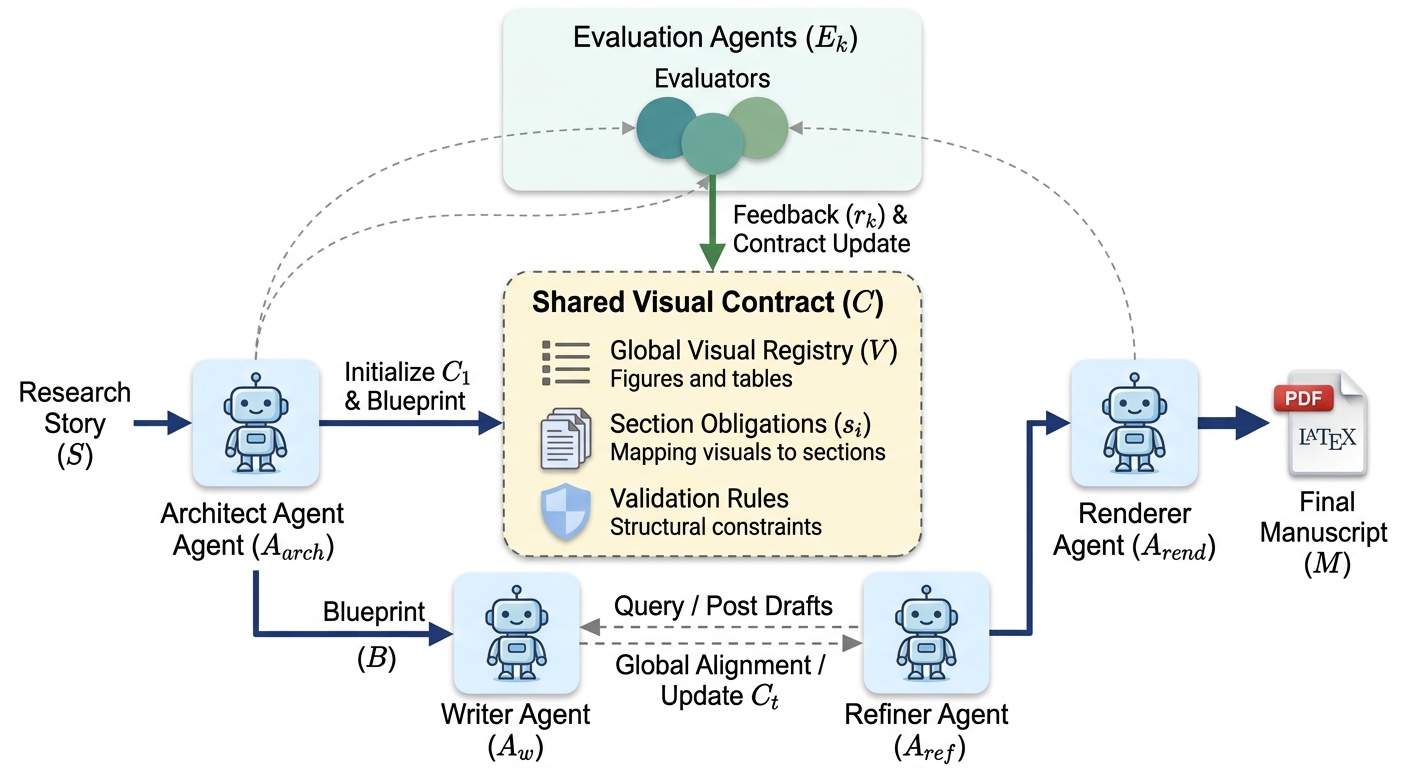}
  \caption{Schema of the shared visual contract used by Story2Proposal. The contract combines a global registry of figures, tables, and citation slots with section-level obligations and validation rules, so that all agents operate on the same structural state and the renderer can deterministically verify label uniqueness, reference resolution, and narrative--visual alignment before compilation.}
  \label{fig:visual_contract}
\end{figure*}

Figure \ref{fig:visual_contract} illustrates the contract schema.

The contract contains three layers of information.

First, a global visual registry maintains the set $V$ of required visual artifacts. Each entry records the artifact type (figure or table), semantic description, canonical label identifier, and expected reference locations within the manuscript.

Second, section-level obligations specify which visual elements must appear within each section $s_i$. These constraints ensure that narrative explanations remain aligned with the figures or tables supporting the scientific claims.

Third, validation rules enforce document-wide consistency requirements such as unique labels, valid cross-references, and alignment between visual descriptions and their textual context.

The contract design is inspired by structured guardrail mechanisms that constrain language model outputs through explicit rule systems and taxonomies \cite{aegis2024aegis2}. In contrast to safety guardrails, however, the contract in Story2Proposal enforces scientific document structure and visual integrity.

Embedding the contract as a persistent shared state allows agents to remain aware of structural obligations during generation. This prevents missing figures, misplaced references, or inconsistent visual descriptions that commonly arise in unconstrained text generation.

\subsection{Multi-Agent Generation Pipeline}

Manuscript generation proceeds through a coordinated pipeline of four agents operating over the shared contract state. The pipeline is illustrated in Figure \ref{fig:generation_pipeline}.


The architect agent transforms the research story $S$ into a structured manuscript blueprint while initializing the contract. It decomposes the narrative into an ordered section structure $\{s_i\}$ and specifies the argument outline for each section, including key claims and supporting evidence. The architect also identifies candidate visual artifacts and registers them in the contract’s visual registry $V$, assigning semantic descriptions and canonical labels. Each artifact is then mapped to section-level obligations indicating where it must appear in the manuscript. Formally, the architect produces a blueprint $B$ and updated contract state:

\begin{equation}
(B, C_1) = A_{arch}(S, C_0)
\end{equation}

The writer agent generates section drafts that realize the blueprint while satisfying contract constraints. Given a section specification $s_i$ and contract state $C_t$, the writer produces a draft $D_i$ that follows the planned argument structure and includes the required visual references registered in the contract. Visual markers and citation identifiers must match the contract registry to maintain consistent references throughout the document. The drafting step can be expressed as

\begin{equation}
D_i = A_w(s_i, C_t)
\end{equation}

The refiner agent performs global alignment over the set of generated drafts $\{D_i\}$. Its role is to improve coherence and consistency across sections while reconciling the manuscript with the contract state. The refiner compresses redundant explanations, harmonizes terminology across sections, and ensures that each visual element referenced in the manuscript is described appropriately in its surrounding text. If inconsistencies are detected, the refiner may trigger contract updates through evaluation feedback. The refinement stage produces a consolidated manuscript:

\begin{equation}
(M', C_{t+1}) = A_{ref}(\{D_i\}, C_t)
\end{equation}

The renderer agent converts the refined manuscript into stable LaTeX output while enforcing deterministic structural validation. During this stage, all visual references are resolved, label identifiers are standardized, and cross-references are validated against the contract. Each artifact in $V$ must appear exactly once and must be referenced consistently within the text. The renderer outputs the final manuscript artifact:

\begin{equation}
M = A_{rend}(M', C_t)
\end{equation}

Separating planning, drafting, refinement, and rendering reduces error propagation and allows each agent to operate with specialized context. The architect establishes the structural skeleton, the writer produces local drafts, the refiner enforces global narrative consistency, and the renderer guarantees structural correctness at the document level.

\subsection{Evaluation and Contract Updates}

Evaluation agents monitor intermediate artifacts during generation and provide feedback signals used to update the contract state. Let $\{E_k\}$ denote the set of evaluation agents, each responsible for a specific dimension such as reasoning verification, data fidelity assessment, or visual consistency.

Given an intermediate artifact $X$ and contract state $C_t$, evaluation agent $E_k$ produces a feedback signal

\begin{equation}
r_k = E_k(X, C_t)
\end{equation}

These signals describe detected issues, confidence estimates, or recommended corrections. Feedback signals are aggregated to update the contract:

\begin{equation}
C_{t+1} = \text{Update}(C_t, \{r_k\})
\end{equation}

Contract updates may introduce additional validation rules, modify visual placement constraints, or require additional explanatory context for specific artifacts. For example, if an evaluation agent detects that a figure reference lacks supporting explanation, the contract may require a descriptive paragraph to accompany the reference.

Embedding evaluation within the generation pipeline enables early detection of inconsistencies and prevents structural errors from propagating through later stages. This generate–evaluate–adapt mechanism reflects emerging architectures for autonomous agent ecosystems, where agents coordinate through shared environments and communication protocols to solve complex tasks collaboratively \cite{yang2025surveyaiagentprotocols}. Such ecosystems support adaptive reasoning and coordinated decision-making across specialized agents \cite{Wang_2024}.

\subsection{Optimization Objective and Algorithm}

Although Story2Proposal does not train a single monolithic model, the evaluation feedback signals can be interpreted as rewards guiding system-level optimization. Let $R(M)$ denote the aggregated evaluation score for a generated manuscript $M$:

\begin{equation}
R(M) = \sum_k w_k r_k
\end{equation}

where $r_k$ represents the feedback from evaluation agent $E_k$ and $w_k$ denotes the relative importance assigned to each evaluation dimension. The generation objective is to maximize $R(M)$ while satisfying all constraints encoded in the contract.

The overall procedure can be summarized as follows:

\begin{algorithm}[htb]
\adjustbox{max width=\columnwidth}{%
\begin{minipage}{\linewidth}
\caption{Story2Proposal Generation Procedure}
\begin{algorithmic}[1]
\STATE Input: research story $S$, initial contract $C_0$, evaluators
\STATE Output: manuscript $M$
\STATE $(B, C) \leftarrow architect(S, C_0)$
\STATE drafts $\leftarrow [\ ]$
\STATE for each section in $B.sections$ do
\STATE $D \leftarrow writer(section, C)$
\STATE feedback $\leftarrow [E(D, C)\ for\ E\ in\ evaluators]$
\STATE $C \leftarrow update\_contract(C, feedback)$
\STATE append $D$ to drafts
\STATE end for
\STATE $(M\_prime, C) \leftarrow refiner(drafts, C)$
\STATE feedback $\leftarrow [E(M\_prime, C)\ for\ E\ in\ evaluators]$
\STATE $C \leftarrow update\_contract(C, feedback)$
\STATE $M \leftarrow renderer(M\_prime, C)$
\STATE return $M$
\end{algorithmic}
\end{minipage}}
\end{algorithm}

The algorithm highlights the core principle of Story2Proposal: manuscript generation emerges from coordinated interactions among specialized agents operating under a shared contract that evolves through evaluation feedback. This mechanism transforms the writing process from a static generation pipeline into an adaptive ecosystem where generation and verification continuously guide one another.

%% file: sections/experiments.tex
\section{Experiments}

\subsection{Implementation Details}

\begin{figure*}[t]
  \centering
  \includegraphics[width=0.8\textwidth]{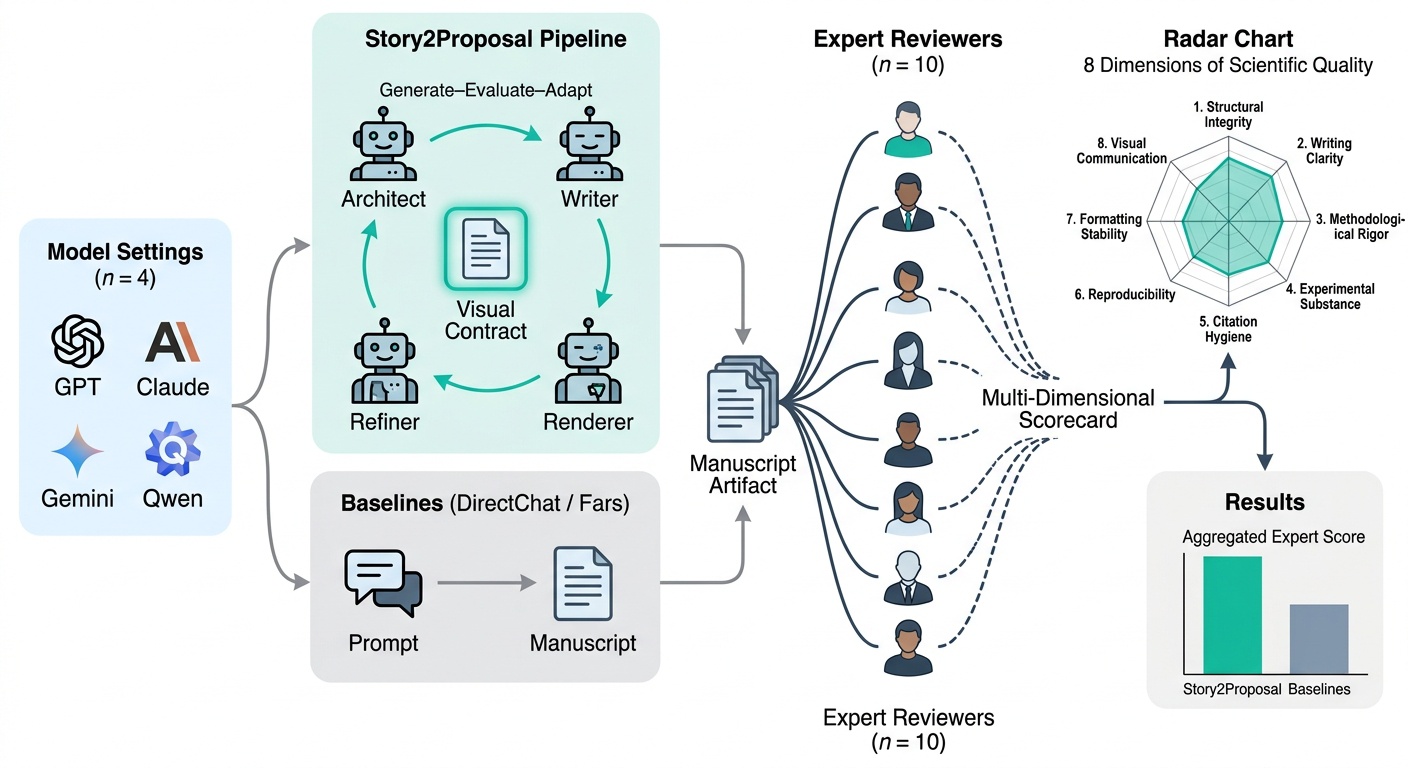}
  \caption{Expert evaluation protocol used in our experiments. Each condition combines one of two generation methods, one of four LLM backbones (GPT, Claude, Gemini, and Qwen), and ten independent expert reviewers. Reviewers assess complete manuscripts on structural integrity, writing clarity, methodological rigor, experimental substance, citation hygiene, reproducibility, formatting stability, and visual communication, and the resulting scores are aggregated into the overall expert evaluation metric.}
  \label{fig:evaluation_protocol}
\end{figure*}

The Story2Proposal system was implemented as a coordinated pipeline of four agents (architect agent, writer agent, refiner agent, and renderer agent) that composed the generation workflow. \cite{zhu2026automatedsafetybenchmarkingmultiagent} Each agent operated as a language-model-driven component interacting through the shared visual contract and the evolving contract state. Agents executed sequentially within the generate–evaluate–adapt loop described in Section 3, where evaluation agents analyzed intermediate artifacts and produced feedback signals that updated the contract state before subsequent generation stages, reflecting emerging practices in evaluating autonomous agent behavior in realistic task environments \cite{zhou2024webarena}. To evaluate robustness across model backbones, the same pipeline configuration was executed using four widely used large language models: GPT, Claude, Gemini, and Qwen. Each run generated a complete manuscript artifact from the same research story input while enforcing contract-governed structural constraints and validating visual artifacts during rendering. The renderer agent produced deterministic LaTeX output and verified compliance with the persistent contract before finalizing the manuscript artifact used for evaluation, and all compared systems were executed with identical prompts and experimental inputs to ensure fairness across model settings and baseline methods in a controlled multi-agent evaluation setup \cite{shu2024genai_multiagent}.

\begin{table*}[t]
\centering
\caption{Average expert evaluation scores comparing Story2Proposal and DirectChat across four LLM settings. Higher is better.}
\label{tab:main_results}
\small
\setlength{\tabcolsep}{14pt}
\resizebox{\textwidth}{!}{%
\begin{tabular}{lccccccccccc}
\toprule

Model & E1 & E2 & E3 & E4 & E5 & E6 & E7 & E8 & E9 & E10 & Avg \\
\midrule
\multicolumn{12}{l}{\textbf{Based on DirectChat}} \\ 

GPT    & 3.62 & 3.01 & 5.97 & 3.33 & 3.49 & 4.06 & 5.34 & 4.23 & 3.64 & 4.09 & 4.08 \\
Claude & 3.39 & 2.68 & 5.98 & 3.04 & 3.36 & 3.96 & 4.88 & 3.92 & 3.23 & 4.20 & 3.86 \\
Gemini & 3.32 & 2.79 & 6.07 & 3.16 & 3.51 & 3.66 & 5.14 & 3.99 & 3.56 & 4.14 & 3.93 \\
Qwen   & 3.38 & 2.74 & 6.19 & 3.22 & 3.41 & 3.93 & 5.32 & 4.05 & 3.58 & 3.93 & 3.98 \\
\midrule

\multicolumn{12}{l}{\textbf{Based on Story2Proposal}} \\
GPT    & 5.39 & 4.99 & 6.98 & 5.03 & 5.18 & 6.25 & 7.13 & 7.60 & 5.09 & 5.98 & 5.96 \\
Claude & 5.70 & 5.53 & 7.22 & 5.35 & 5.30 & 6.01 & 6.83 & 7.81 & 5.58 & 6.20 & 6.15 \\
Gemini & 5.60 & 5.61 & 7.42 & 5.51 & 5.59 & 6.17 & 7.23 & 7.54 & 5.52 & 6.38 & 6.26 \\
Qwen   & 6.07 & 5.60 & 7.24 & 5.35 & 5.44 & 6.13 & 7.09 & 7.20 & 5.83 & 6.12 & 6.21 \\
\bottomrule
\end{tabular}%
} 
\end{table*}

\subsection{Experimental Design}

The evaluation protocol assessed whether contract-governed generation improved the robustness and structural reliability of automatically generated research manuscripts. \cite{lin2025factaudit} The benchmark environment used the Jericho research corpus to construct research stories and manuscript tasks, enabling evaluation of complex reasoning and structured document synthesis in a controlled setting. Human expert reviewers served as the primary evaluation mechanism because automated metrics often fail to capture reasoning coherence and document-level scientific quality in generated manuscripts \cite{golovneva2023roscoe}. Prior work has similarly emphasized the importance of human-grounded evaluation when assessing complex agent systems and generation pipelines \cite{ren2024bases}. The evaluation followed a 2 $\times$ 4 $\times$ 10 design consisting of two generation methods, four model settings, and ten independent expert reviewers per condition. Reviewers evaluated manuscripts across eight dimensions of scientific writing quality: structural integrity, writing clarity, methodological rigor, experimental substance, citation hygiene, reproducibility, formatting stability, and visual communication. Scores across these dimensions were aggregated into a single expert evaluation score for each generated manuscript. Figure \ref{fig:evaluation_protocol} illustrates the full evaluation design including model configurations and reviewer structure. The first experiment compared Story2Proposal with DirectChat, a single-stage prompt-based manuscript generation baseline without structural contracts or multi-agent coordination. This comparison tested whether contract-governed generation improved document-level organization and reasoning coherence relative to direct language model generation. The second experiment compared Story2Proposal with Fars, a structured manuscript generation baseline evaluated on four benchmark research papers. This comparison examined whether the contract-centered architecture provided improvements even against a structured system designed for research manuscript synthesis. The protocol follows established practices for evaluating complex multi-agent workflows using structured output analysis and systematic scoring procedures \cite{agentcompass2024evaluation}.

\subsection{Results}

Table \ref{tab:main_results} reports the primary comparison between Story2Proposal and DirectChat across four model settings. Story2Proposal outperformed DirectChat in every configuration, indicating that contract-governed generation improved manuscript quality independently of the underlying language model. With GPT, Story2Proposal achieved an expert evaluation score of 5.962 compared with 4.078 for DirectChat. Claude produced 6.153 versus 3.864, Gemini produced 6.257 versus 3.934, and Qwen produced 6.207 versus 3.975. Averaged across all model settings, Story2Proposal achieved 6.145 compared with 3.963 for DirectChat, an improvement of +2.182. These results indicated that the contract-governed multi-agent framework consistently produced higher-quality manuscripts across diverse model backbones. Expert reviewers reported fewer structural inconsistencies, clearer experimental descriptions, and more reliable formatting in Story2Proposal outputs. Improvements were particularly visible in structural integrity and formatting stability, where manuscripts generated through the contract maintained more consistent section organization and fewer layout errors than those produced by DirectChat.

\begin{table*}[t]
\centering
\caption{Average expert evaluation scores comparing Story2Proposal and Fars across four benchmark research papers. Higher is better.}
\label{tab:secondary_results}
\setlength{\tabcolsep}{14pt}
\resizebox{\textwidth}{!}{%
\begin{tabular}{lccccccccccc}
\toprule
Paper & E1 & E2 & E3 & E4 & E5 & E6 & E7 & E8 & E9 & E10 & Avg \\
\midrule
\multicolumn{12}{l}{\textbf{Based on Fars}} \\
Escrowed Batch Reveal & 4.55 & 4.39 & 6.06 & 4.41 & 4.38 & 5.04 & 6.19 & 4.57 & 4.47 & 5.22 & 4.93 \\
Symbolic Execution & 4.97 & 5.64 & 6.32 & 5.40 & 5.65 & 5.57 & 6.46 & 5.11 & 5.57 & 5.69 & 5.64 \\
Hazard-Signature Tombstones & 4.27 & 4.06 & 6.09 & 4.19 & 4.33 & 4.57 & 6.04 & 4.59 & 4.21 & 4.74 & 4.71 \\
Poisoning LLM-Induced Rules & 5.00 & 5.59 & 6.01 & 5.11 & 5.35 & 5.22 & 6.80 & 5.26 & 5.14 & 5.63 & 5.41 \\
\midrule
\multicolumn{12}{l}{\textbf{Based on Story2Proposal}} \\
Escrowed Batch Reveal & 5.72 & 4.49 & 6.61 & 4.62 & 4.67 & 5.27 & 6.94 & 5.87 & 4.87 & 5.53 & 5.46 \\
Symbolic Execution & 5.66 & 5.20 & 6.79 & 5.16 & 5.34 & 5.67 & 7.54 & 5.90 & 5.28 & 6.32 & 5.89 \\
Hazard-Signature Tombstones & 4.96 & 4.87 & 6.52 & 4.74 & 5.24 & 5.38 & 7.00 & 6.00 & 5.29 & 6.00 & 5.60 \\
Poisoning LLM-Induced Rules & 5.61 & 5.22 & 6.68 & 4.79 & 5.56 & 5.88 & 7.15 & 6.38 & 5.32 & 6.17 & 5.88 \\
\bottomrule
\end{tabular}%
}

\end{table*}

Table \ref{tab:secondary_results} presents the benchmark comparison between Story2Proposal and Fars across four research papers: Escrowed Batch Reveal, Symbolic Execution, Hazard-Signature Tombstones, and Poisoning LLM-Induced Rules. Story2Proposal achieved higher expert evaluation scores on all four papers. For Escrowed Batch Reveal, the score increased from 4.928 with Fars to 5.459 with Story2Proposal. For Symbolic Execution, the score increased from 5.638 to 5.886. Hazard-Signature Tombstones improved from 4.709 to 5.600, and Poisoning LLM-Induced Rules improved from 5.511 to 5.876. Averaged across the four benchmark papers, Story2Proposal achieved a mean expert evaluation score of 5.705 compared with 5.197 for Fars, corresponding to an improvement of +0.508. Although the margin was smaller than in the DirectChat comparison, the results indicated that maintaining an evolving contract state and enforcing structural obligations during generation improved document reliability even relative to a structured baseline. Expert reviewers also reported stronger alignment between narrative explanations and visual artifacts, particularly in experimental sections where figures and tables were consistently referenced and correctly positioned. No ablation study was included in this version of the system evaluation.

%% file: sections/analysis.tex
\section{Analysis}

\subsection{Structural Robustness and Feedback Mechanisms}

\begin{figure}[t]
  \centering
  \includegraphics[width=0.5\textwidth]{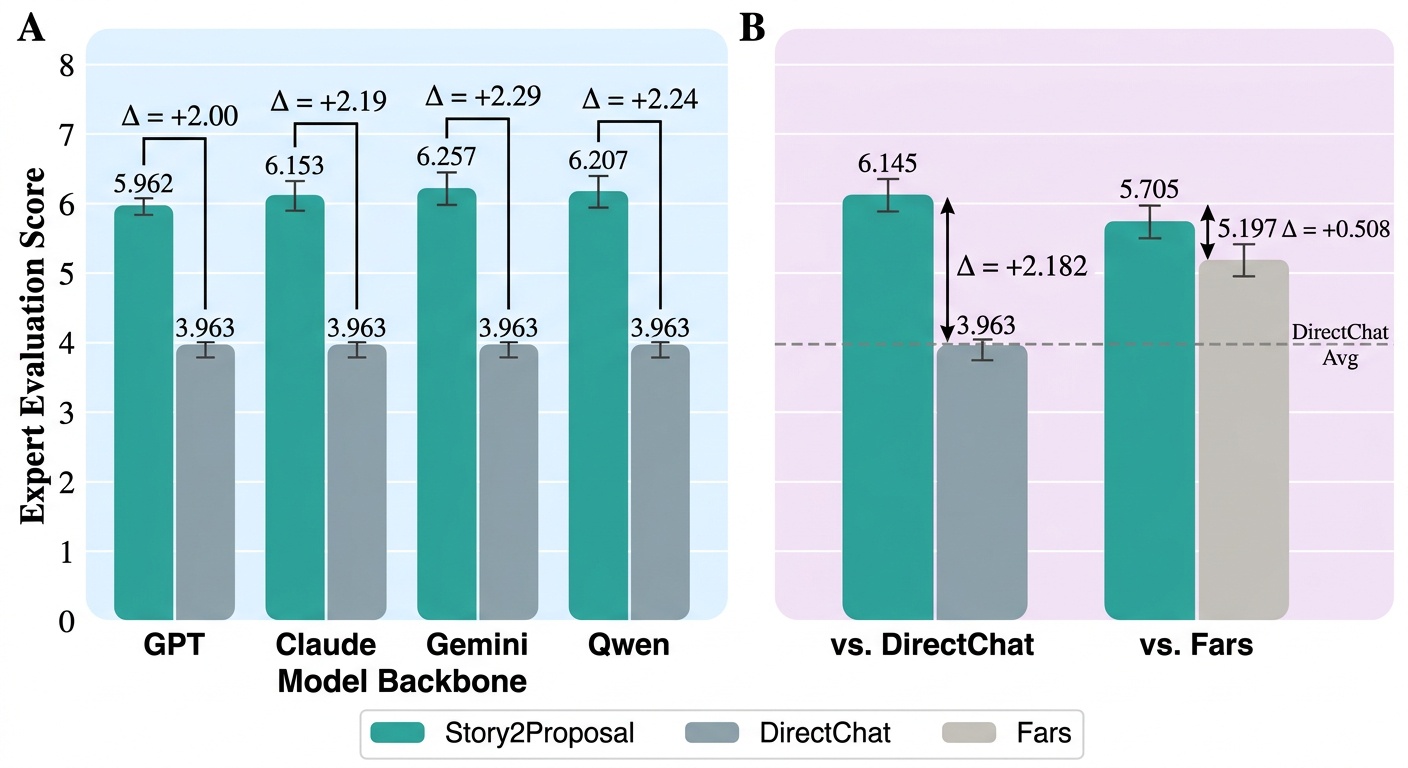}
  \caption{Average performance deltas between Story2Proposal and baseline generation methods across experiments. Positive values indicate that the contract-governed multi-agent framework outperforms the comparison system under the same backbone or benchmark setting. The figure summarizes both the large gains over DirectChat in the cross-model study and the smaller but consistent improvements over the structured Fars baseline, highlighting the benefit of stronger structural control and narrative--visual alignment.}
  \label{fig:delta_comparison}
\end{figure}

The primary improvement observed for Story2Proposal arose from enforcing structural constraints through the persistent shared visual contract during manuscript generation, a strategy consistent with prior work showing that explicit planning and structured narrative scaffolds can improve coherence and organization in generated outputs \cite{li2025storytellerenhancedplotplanningframework}. As illustrated in Figure~\ref{fig:delta_comparison}, manuscripts produced by Story2Proposal consistently achieved higher expert evaluation scores than those generated by DirectChat across all evaluated model backbones.  The system reached an average expert evaluation score of 6.145 compared with 3.963 for DirectChat, corresponding to an improvement of +2.182. Because this improvement appeared consistently across different model backbones, the results indicate that the advantage derived primarily from the contract-governed architecture rather than from any specific language model capability, aligning with broader evidence that structured multi-component or collaborative reasoning systems can outperform single-agent generation pipelines on complex tasks \cite{twoheads2024multiagent}.

A central mechanism behind this improvement is the contract state maintained throughout the generation process. The persistent shared visual contract specifies section structure, visual artifacts, and validation rules that must be satisfied during manuscript construction. By embedding these constraints directly into generation, Story2Proposal prevents common failure modes of unconstrained generation pipelines such as structural drift, missing visual artifacts, or inconsistent section organization. \cite{article} Agents therefore operate within an explicit structural specification rather than relying on implicit formatting decisions during text generation.

This behavior aligns with prior work showing that structured constraints can improve reliability in large language model generation. Grammar-guided generation systems demonstrate that restricting output structure enables earlier detection and correction of inconsistencies during generation rather than after completion \cite{ugare2025itergeniterativesemanticawarestructured}. Story2Proposal applies a similar principle at the document level: each stage of the pipeline updates and validates the contract state before subsequent generation steps proceed, preventing structural errors from accumulating across sections.

An additional contributor to robustness is the generate--evaluate--adapt loop that integrates evaluation agents into the generation process. Instead of performing evaluation only after the manuscript is produced, intermediate artifacts are continuously assessed for reasoning quality, structural integrity, and visual consistency. Feedback signals from evaluation agents modify the contract state, enabling the system to correct issues early in the pipeline.

This feedback-driven process resembles interpretable reasoning frameworks that expose intermediate reasoning stages so that errors can be detected before final outputs are produced. Such approaches improve controllability and transparency by making intermediate reasoning states observable and correctable \cite{chen-etal-2025-cod}. In Story2Proposal, evaluation agents play an analogous role by monitoring intermediate manuscript states and triggering corrective updates when inconsistencies are detected.

\subsection{Cross-Model Consistency and Visual Communication Effects}

A key empirical observation is that the performance improvement remained stable across multiple language model backbones. Story2Proposal consistently outperformed DirectChat for GPT, Claude, Gemini, and Qwen. The system achieved expert evaluation scores of 5.962 for GPT, 6.153 for Claude, 6.257 for Gemini, and 6.207 for Qwen, while DirectChat produced substantially lower scores across the same models. This pattern suggests that the architectural structure of Story2Proposal, rather than model-specific reasoning ability, primarily explains the observed improvements.

In unconstrained generation pipelines such as DirectChat, the language model must simultaneously manage reasoning, formatting, section structure, and visual references. This increases the complexity of the generation task and often leads to structural inconsistencies. Story2Proposal externalizes these responsibilities through the persistent shared visual contract and specialized agents, allowing the language model to focus on localized reasoning and content generation while structural constraints are handled by the system architecture.

The benefits of such decomposition are consistent with findings from multi-agent simulation environments where distributing responsibilities across coordinated agents improves reliability and stability in complex workflows \cite{ren-etal-2024-bases}. By separating planning, drafting, refinement, and rendering responsibilities, Story2Proposal maintains consistent structural behavior even when the underlying language model changes.

The visual contract also directly improved visual communication and formatting stability. In prompt-based generation pipelines, figures and tables are often lost or duplicated during iterative rewriting because they are represented only as text tokens. Story2Proposal instead registers visual artifacts in a contract-based registry with explicit identifiers and placement obligations. Agents reference these registered artifacts rather than regenerating figure markers, ensuring that visual elements remain aligned with the narrative throughout the document.

This design difference is reflected in the comparison with the Fars baseline. Across the four benchmark papers---Escrowed Batch Reveal, Symbolic Execution, Hazard-Signature Tombstones, and Poisoning LLM-Induced Rules---Story2Proposal achieved higher expert evaluation scores than Fars in every case, with an average score of 5.705 compared with 5.197 for Fars, corresponding to an improvement of +0.508. Because Fars already employs structured templates, the remaining improvement suggests that persistent contract enforcement and iterative validation further strengthen alignment between narrative explanations and visual artifacts.

Prior studies on visual communication systems also highlight that maintaining explicit alignment between narrative text and visual transformations improves interpretability in complex documents \cite{Hong2023VisualTT}. By encoding such alignment constraints directly within the contract state, Story2Proposal preserves the semantic relationship between visual artifacts and their textual explanations during generation.

\subsection{Failure Mode Analysis}

Despite the consistent gains observed in the experiments, the results also reveal limitations of the contract-governed generation approach. The improvement over the Fars baseline was smaller than the improvement over DirectChat. While Story2Proposal outperformed Fars across all evaluated benchmark papers, the margin of improvement was notably reduced. This pattern indicates that the largest benefits arise when replacing unconstrained prompt-based generation systems, whereas structured baselines already mitigate many basic structural errors.

Another limitation arises from the reliance on evaluation agents for detecting inconsistencies. If evaluation agents fail to identify subtle reasoning weaknesses---such as an interpretation that satisfies structural requirements but remains scientifically weak---the contract state may not trigger corrective updates. In such cases, the generated manuscript can remain structurally valid while containing less rigorous reasoning.

Improving this stage may require richer evaluation signals and more specialized critique modules. Structured risk taxonomies and evaluation frameworks have been shown to improve the identification of nuanced categories in language model outputs \cite{aegis2024aegis2}, suggesting that more sophisticated evaluation agents could strengthen the feedback mechanism in future systems.

A final limitation concerns incomplete research inputs. Because Story2Proposal enforces structural and visual obligations rather than generating missing scientific evidence, manuscripts derived from incomplete research stories may remain logically consistent yet still reflect gaps in the underlying narrative. Addressing this issue would likely require integrating retrieval mechanisms or evidence verification systems capable of expanding or validating the input research story before manuscript generation begins.

%% file: sections/conclusion.tex
\section{Conclusion}

This work presented Story2Proposal, a contract-governed multi-agent framework that models automated scientific manuscript generation as a structured construction process in which specialized agents transform a research story into a coherent manuscript while enforcing shared structural and visual obligations. Empirical results showed consistent improvements over unconstrained generation pipelines: across four model backbones, Story2Proposal achieved expert evaluation scores of 5.962 (GPT), 6.153 (Claude), 6.257 (Gemini), and 6.207 (Qwen), while DirectChat produced 4.078, 3.864, 3.934, and 3.975, yielding an overall score of 6.145 versus 3.963 and an improvement of +2.182 points. Comparisons with the structured baseline Fars also demonstrated gains in document reliability and alignment between narrative text and visual artifacts, with Story2Proposal averaging 5.705 versus 5.197, including improvements on Escrowed Batch Reveal (5.459 vs 4.928), Symbolic Execution (5.886 vs 5.638), Hazard-Signature Tombstones (5.600 vs 4.709), and Poisoning LLM-Induced Rules (5.876 vs 5.511). These findings indicate that maintaining a persistent shared visual contract and evolving contract state helps preserve structural integrity, formatting stability, and visual communication during generation, suggesting that architectural constraints play a central role in reliable manuscript construction. Improvements over template-based systems such as Fars remain smaller, and performance still depends on the completeness of the input research story and the ability of evaluation agents to detect subtle reasoning issues. Future work may extend contract-centered generation with stronger verification and domain-specific validation mechanisms for reliable AI-assisted scientific communication while supporting interpretable reasoning in automated analysis systems.

%% file: paper.bib
@article{agashe2024agent,
  title={Agent s: An open agentic framework that uses computers like a human},
  author={Agashe, Saaket and Han, Jiuzhou and Gan, Shuyu and Yang, Jiachen and Li, Ang and Wang, Xin Eric},
  journal={arXiv preprint arXiv:2410.08164},
  year={2024}
}

@article{feng2022training,
  title={Training-free structured diffusion guidance for compositional text-to-image synthesis},
  author={Feng, Weixi and He, Xuehai and Fu, Tsu-Jui and Jampani, Varun and Akula, Arjun and Narayana, Pradyumna and Basu, Sugato and Wang, Xin Eric and Wang, William Yang},
  journal={arXiv preprint arXiv:2212.05032},
  year={2022}
}

@article{xu2026idea2paper,
  title={Idea2Paper: What Should an End-to-End Research Agent Really Do?},
  author={Xu, Tengyue and Qian, Zhuoyang and Liu, Gaoge and Zhang, Zhentao and Ling, Li and Wu, Biao and Zhang, Shuo and Lu, Ke and Shi, Wei and Wang, Ziqi and others},
  year={2026}
}

@article{khot2022decomposed,
  title={Decomposed prompting: A modular approach for solving complex tasks},
  author={Khot, Tushar and Trivedi, Harsh and Finlayson, Matthew and Fu, Yao and Richardson, Kyle and Clark, Peter and Sabharwal, Ashish},
  journal={arXiv preprint arXiv:2210.02406},
  year={2022}
}

@article{moskovitz2023confronting,
  title={Confronting reward model overoptimization with constrained RLHF},
  author={Moskovitz, Ted and Singh, Aaditya K and Strouse, DJ and Sandholm, Tuomas and Salakhutdinov, Ruslan and Dragan, Anca D and McAleer, Stephen},
  journal={arXiv preprint arXiv:2310.04373},
  year={2023}
}

@article{shang2024agentsquare,
  title={Agentsquare: Automatic llm agent search in modular design space},
  author={Shang, Yu and Li, Yu and Zhao, Keyu and Ma, Likai and Liu, Jiahe and Xu, Fengli and Li, Yong},
  journal={arXiv preprint arXiv:2410.06153},
  year={2024}
}

@article{zeng2022socratic,
  title={Socratic models: Composing zero-shot multimodal reasoning with language},
  author={Zeng, Andy and Attarian, Maria and Ichter, Brian and Choromanski, Krzysztof and Wong, Adrian and Welker, Stefan and Tombari, Federico and Purohit, Aveek and Ryoo, Michael and Sindhwani, Vikas and others},
  journal={arXiv preprint arXiv:2204.00598},
  year={2022}
}

@article{zhang2022automatic,
  title={Automatic chain of thought prompting in large language models},
  author={Zhang, Zhuosheng and Zhang, Aston and Li, Mu and Smola, Alex},
  journal={arXiv preprint arXiv:2210.03493},
  year={2022}
}

@article{zheng2023take,
  title={Take a step back: Evoking reasoning via abstraction in large language models},
  author={Zheng, Huaixiu Steven and Mishra, Swaroop and Chen, Xinyun and Cheng, Heng-Tze and Chi, Ed H and Le, Quoc V and Zhou, Denny},
  journal={arXiv preprint arXiv:2310.06117},
  year={2023}
}

@article{lu2024ai,
  title={The ai scientist: Towards fully automated open-ended scientific discovery},
  author={Lu, Chris and Lu, Cong and Lange, Robert Tjarko and Foerster, Jakob and Clune, Jeff and Ha, David},
  journal={arXiv preprint arXiv:2408.06292},
  year={2024}
}

@article{gottweis2025towards,
  title={Towards an AI co-scientist},
  author={Gottweis, Juraj and Weng, Wei-Hung and Daryin, Alexander and Tu, Tao and Palepu, Anil and Sirkovic, Petar and Myaskovsky, Artiom and Weissenberger, Felix and Rong, Keran and Tanno, Ryutaro and others},
  journal={arXiv preprint arXiv:2502.18864},
  year={2025}
}

@article{weng2025deepscientist,
  title={Deepscientist: Advancing frontier-pushing scientific findings progressively},
  author={Weng, Yixuan and Zhu, Minjun and Xie, Qiujie and Sun, Qiyao and Lin, Zhen and Liu, Sifan and Zhang, Yue},
  journal={arXiv preprint arXiv:2509.26603},
  year={2025}
}

@article{gridach2025agentic,
  title={Agentic ai for scientific discovery: A survey of progress, challenges, and future directions},
  author={Gridach, Mourad and Nanavati, Jay and Abidine, Khaldoun Zine El and Mendes, Lenon and Mack, Christina},
  journal={arXiv preprint arXiv:2503.08979},
  year={2025}
}

@article{chen2025ai4research,
  title={Ai4research: A survey of artificial intelligence for scientific research},
  author={Chen, Qiguang and Yang, Mingda and Qin, Libo and Liu, Jinhao and Yan, Zheng and Guan, Jiannan and Peng, Dengyun and Ji, Yiyan and Li, Hanjing and Hu, Mengkang and others},
  journal={arXiv preprint arXiv:2507.01903},
  year={2025}
}

@article{zhu2026autofigure,
  title={AutoFigure: Generating and Refining Publication-Ready Scientific Illustrations},
  author={Zhu, Minjun and Lin, Zhen and Weng, Yixuan and Lu, Panzhong and Xie, Qiujie and Wei, Yifan and Liu, Sifan and Sun, Qiyao and Zhang, Yue},
  journal={arXiv preprint arXiv:2602.03828},
  year={2026}
}

@article{weng2024cycleresearcher,
  title={Cycleresearcher: Improving automated research via automated review},
  author={Weng, Yixuan and Zhu, Minjun and Bao, Guangsheng and Zhang, Hongbo and Wang, Jindong and Zhang, Yue and Yang, Linyi},
  journal={arXiv preprint arXiv:2411.00816},
  year={2024}
}

@article{m2024augmenting,
  title={Augmenting large language models with chemistry tools},
  author={M. Bran, Andres and Cox, Sam and Schilter, Oliver and Baldassari, Carlo and White, Andrew D and Schwaller, Philippe},
  journal={Nature machine intelligence},
  volume={6},
  number={5},
  pages={525--535},
  year={2024},
  publisher={Nature Publishing Group UK London}
}

@article{boiko2023autonomous,
  title={Autonomous chemical research with large language models},
  author={Boiko, Daniil A and MacKnight, Robert and Kline, Ben and Gomes, Gabe},
  journal={Nature},
  volume={624},
  number={7992},
  pages={570--578},
  year={2023},
  publisher={Nature Publishing Group UK London}
}

@article{tang2025ai,
  title={Ai-researcher: Autonomous scientific innovation},
  author={Tang, Jiabin and Xia, Lianghao and Li, Zhonghang and Huang, Chao},
  journal={arXiv preprint arXiv:2505.18705},
  year={2025}
}

@article{luo2025llm4sr,
  title={Llm4sr: A survey on large language models for scientific research},
  author={Luo, Ziming and Yang, Zonglin and Xu, Zexin and Yang, Wei and Du, Xinya},
  journal={arXiv preprint arXiv:2501.04306},
  year={2025}
}

@inproceedings{zheng2025automation,
  title={From automation to autonomy: A survey on large language models in scientific discovery},
  author={Zheng, Tianshi and Deng, Zheye and Tsang, Hong Ting and Wang, Weiqi and Bai, Jiaxin and Wang, Zihao and Song, Yangqiu},
  booktitle={Proceedings of the 2025 Conference on Empirical Methods in Natural Language Processing},
  pages={17744--17761},
  year={2025}
}

@article{eger2025transforming,
  title={Transforming science with large language models: A survey on ai-assisted scientific discovery, experimentation, content generation, and evaluation},
  author={Eger, Steffen and Cao, Yong and D'Souza, Jennifer and Geiger, Andreas and Greisinger, Christian and Gross, Stephanie and Hou, Yufang and Krenn, Brigitte and Lauscher, Anne and Li, Yizhi and others},
  journal={arXiv preprint arXiv:2502.05151},
  year={2025}
}

@article{wang2024autosurvey,
  title={AutoSurvey: Large Language Models Can Automatically Write Surveys},
  author={Wang, Yidong and Guo, Qi and Yao, Wenjin and Zhang, Hongbo and Zhang, Xin and Wu, Zhen and Zhang, Meishan and Dai, Xinyu and Zhang, Min and Wen, Qingsong and Ye, Wei and Zhang, Shikun and Zhang, Yue},
  journal={Advances in Neural Information Processing Systems},
  volume={37},
  pages={115119--115145},
  year={2024}
}

@inproceedings{shao2024assisting,
  title={Assisting in Writing Wikipedia-like Articles From Scratch with Large Language Models},
  author={Shao, Yijia and Jiang, Yucheng and Kanell, Theodore A and Xu, Peter and Khattab, Omar and Lam, Monica S},
  booktitle={Proceedings of the 2024 Conference of the North American Chapter of the Association for Computational Linguistics},
  pages={6252--6278},
  year={2024}
}

@article{liang2025surveyx,
  title={SurveyX: Academic Survey Automation via Large Language Models},
  author={Liang, Xun and Yang, Jiawei and Wang, Yezhaohui and Tang, Chao and Zheng, Zihao and Song, Siyu and Lin, Zhangming and Yang, Yue and Niu, Simin and Wang, Hai and Tang, Bo and Xiong, Feiyu and Mao, Keming and Li, Zhiyu},
  journal={arXiv preprint arXiv:2502.14776},
  year={2025}
}

@article{susnjak2025automating,
  title={Automating Research Synthesis with Domain-specific Large Language Model Fine-tuning},
  author={Susnjak, Teo and Hwang, Peter and Reyes, Napoleon and Barczak, Andre L C and McIntosh, Timothy and Ranathunga, Surangika},
  journal={ACM Transactions on Knowledge Discovery from Data},
  volume={19},
  number={3},
  year={2025}
}

@inproceedings{koncel2019text,
  title={Text Generation from Knowledge Graphs with Graph Transformers},
  author={Koncel-Kedziorski, Rik and Bekal, Dhanush and Luan, Yi and Lapata, Mirella and Hajishirzi, Hannaneh},
  booktitle={Proceedings of the 2019 Conference of the North American Chapter of the Association for Computational Linguistics},
  pages={2284--2293},
  year={2019}
}

@inproceedings{yang2023doc,
  title={{DOC}: Improving Long Story Coherence With Detailed Outline Control},
  author={Yang, Kevin and Klein, Dan and Peng, Nanyun and Tian, Yuandong},
  booktitle={Proceedings of the 61st Annual Meeting of the Association for Computational Linguistics},
  pages={3378--3465},
  year={2023}
}

@article{bai2024longwriter,
  title={LongWriter: Unleashing 10,000+ Word Generation from Long Context {LLM}s},
  author={Bai, Yushi and Lv, Jiajie and Song, Hao and Hou, Kaijie and Wang, Ji-Rong and Dong, Yuxiao and Li, Jie and Tang, Jie},
  journal={arXiv preprint arXiv:2408.07055},
  year={2024}
}

@article{huot2024agents,
  title={Agents' Room: Narrative Generation through Multi-step Collaboration},
  author={Huot, Fantine and Amplayo, Reinald Kim and Palomaki, Jennimaria and Jakobovits, Alice Shoshana and Clark, Elizabeth and Lapata, Mirella},
  journal={arXiv preprint arXiv:2410.02603},
  year={2024}
}

@article{wan2025cognitive,
  title={A Cognitive Writing Perspective for Constrained Long-Form Text Generation},
  author={Wan, Kaiyang and Mu, Honglin and Hao, Rui and Luo, Haoran and Gu, Tianle and Chen, Xiuying},
  journal={arXiv preprint arXiv:2502.12568},
  year={2025}
}

@inproceedings{yang2022re3,
  title={{Re3}: Generating Longer Stories with Recursive Reprompting and Revision},
  author={Yang, Kevin and Tian, Yuandong and Peng, Nanyun and Klein, Dan},
  booktitle={Proceedings of the 2022 Conference on Empirical Methods in Natural Language Processing},
  pages={4393--4479},
  year={2022}
}

@article{kim2024navigating,
  title={Navigating the Path of Writing: Outline-guided Text Generation with Large Language Models},
  author={Kim, Yukyung and Lee, Kunyoung and Kim, Seongjin and Lee, Yun-Gyung and Kim, Sungdong and Yeo, Jungwoo},
  journal={arXiv preprint arXiv:2404.13919},
  year={2024}
}

@article{liang2024integrating,
  title={Integrating planning into single-turn long-form text generation},
  author={Liang, Yi and Wu, You and Zhuang, Honglei and Chen, Li and Shen, Jiaming and Jia, Yiling and Qin, Zhen and Sanghai, Sumit and Wang, Xuanhui and Yang, Carl and others},
  journal={arXiv preprint arXiv:2410.06203},
  year={2024}
}

@article{wang2024dome,
  title={Generating Long-form Story Using Dynamic Hierarchical Outlining with Memory-Enhancement},
  author={Wang, Qianyue and Hu, Jinwu and Li, Zhengping and Wang, Yufeng and Li, Daiyuan and Hu, Yu and Tan, Mingkui},
  journal={arXiv preprint arXiv:2412.13575},
  year={2024}
}

@inproceedings{xiong2025beyond,
  title={Beyond outlining: Heterogeneous recursive planning for adaptive long-form writing with language models},
  author={Xiong, Ruibin and Chen, Yimeng and Khizbullin, Dmitrii and Zhuge, Mingchen and Schmidhuber, J{\"u}rgen},
  booktitle={Proceedings of the 2025 Conference on Empirical Methods in Natural Language Processing},
  pages={24689--24725},
  year={2025}
}

@article{wu2023autogen,
  title={Auto{G}en: Enabling Next-Gen {LLM} Applications via Multi-Agent Conversation Framework},
  author={Wu, Qingyun and Bansal, Gagan and Zhang, Jieyu and Wu, Yiran and Zhang, Shaokun and Zhu, Erkang and Li, Beibin and Jiang, Li and Zhang, Xiaoyun and Wang, Chi},
  journal={arXiv preprint arXiv:2308.08155},
  year={2023}
}

@article{schmidgall2025agent,
  title={Agent Laboratory: Using {LLM} Agents as Research Assistants},
  author={Schmidgall, Samuel and Su, Yusheng and Wang, Ze and Sun, Ximeng and Wu, Jialian and Yu, Xiaodong and Liu, Jiang and Liu, Zicheng and Barsoum, Emad},
  journal={arXiv preprint arXiv:2501.04227},
  year={2025}
}

@inproceedings{dathathri2020plug,
  title={Plug and Play Language Models: A Simple Approach to Controlled Text Generation},
  author={Dathathri, Sumanth and Madotto, Andrea and Lan, Janice and Hung, Jane and Frank, Eric and Molino, Piero and Yosinski, Jason and Liu, Rosanne},
  booktitle={International Conference on Learning Representations},
  year={2020}
}

@inproceedings{yang2021fudge,
  title={{FUDGE}: Controlled Text Generation With Future Discriminators},
  author={Yang, Kevin and Klein, Dan},
  booktitle={Proceedings of the 2021 Conference of the North American Chapter of the Association for Computational Linguistics},
  pages={3511--3535},
  year={2021}
}

@article{mudgal2024controlled,
  title={Controlled Decoding from Language Models},
  author={Mudgal, Sidharth and Lee, Jong and Ganapathy, Harish and Li, YaGuang and Wang, Tao and Huang, Yanping and Chen, Zhifeng and Cheng, Heng-Tze and Collins, Michael and Strope, Trevor and others},
  journal={arXiv preprint arXiv:2310.17022},
  year={2024}
}

@article{liu2024improving,
  title={Improving long-text alignment for text-to-image diffusion models},
  author={Liu, Luping and Du, Chao and Pang, Tianyu and Wang, Zehan and Li, Chongxuan and Xu, Dong},
  journal={arXiv preprint arXiv:2410.11817},
  year={2024}
}

@article{mitchener2025kosmos,
  title={Kosmos: An ai scientist for autonomous discovery},
  author={Mitchener, Ludovico and Yiu, Angela and Chang, Benjamin and Bourdenx, Mathieu and Nadolski, Tyler and Sulovari, Arvis and Landsness, Eric C and Barabasi, Daniel L and Narayanan, Siddharth and Evans, Nicky and others},
  journal={arXiv preprint arXiv:2511.02824},
  year={2025}
}

@article{ren2025towards,
  title={Towards scientific intelligence: A survey of llm-based scientific agents},
  author={Ren, Shuo and Xie, Can and Jian, Pu and Ren, Zhenjiang and Leng, Chunlin and Zhang, Jiajun},
  journal={arXiv preprint arXiv:2503.24047},
  year={2025}
}

@article{yamada2025ai,
  title={The ai scientist-v2: Workshop-level automated scientific discovery via agentic tree search},
  author={Yamada, Yutaro and Lange, Robert Tjarko and Lu, Cong and Hu, Shengran and Lu, Chris and Foerster, Jakob and Clune, Jeff and Ha, David},
  journal={arXiv preprint arXiv:2504.08066},
  year={2025}
}

@article{li2025agentswift,
  title={Agentswift: Efficient llm agent design via value-guided hierarchical search},
  author={Li, Yu and Li, Lehui and Wu, Zhihao and Liao, Qingmin and Hao, Jianye and Shao, Kun and Xu, Fengli and Li, Yong},
  journal={arXiv preprint arXiv:2506.06017},
  year={2025}
}

@article{xi2025survey,
  title={A survey of llm-based deep search agents: Paradigm, optimization, evaluation, and challenges},
  author={Xi, Yunjia and Lin, Jianghao and Xiao, Yongzhao and Zhou, Zheli and Shan, Rong and Gao, Te and Zhu, Jiachen and Liu, Weiwen and Yu, Yong and Zhang, Weinan},
  journal={arXiv preprint arXiv:2508.05668},
  year={2025}
}

@article{shi2025deep,
  title={Deep research: A systematic survey},
  author={Shi, Zhengliang and Chen, Yiqun and Li, Haitao and Sun, Weiwei and Ni, Shiyu and Lyu, Yougang and Fan, Run-Ze and Jin, Bowen and Weng, Yixuan and Zhu, Minjun and others},
  journal={arXiv preprint arXiv:2512.02038},
  year={2025}
}

@inproceedings{hang2025beyond,
  title={Beyond Search: Measuring LLM Performance for Scientific Literature Discovery},
  author={Hang, Ching Nam and Yu, Pei-Duo and Tan, Chee Wei and Chiu, Dah Ming},
  booktitle={2025 IEEE International Conference on Teaching, Assessment, and Learning for Engineering (TALE)},
  pages={1--7},
  year={2025},
  organization={IEEE}
}

@article{huang2025deep,
  title={Deep research agents: A systematic examination and roadmap},
  author={Huang, Yuxuan and Chen, Yihang and Zhang, Haozheng and Li, Kang and Zhou, Huichi and Fang, Meng and Yang, Linyi and Li, Xiaoguang and Shang, Lifeng and Xu, Songcen and others},
  journal={arXiv preprint arXiv:2506.18096},
  year={2025}
}

@article{luo2025dr,
  title={Dr. V: A Hierarchical Perception-Temporal-Cognition Framework to Diagnose Video Hallucination by Fine-grained Spatial-Temporal Grounding},
  author={Luo, Meng and Wu, Shengqiong and Jing, Liqiang and Ju, Tianjie and Zheng, Li and Lai, Jinxiang and Wu, Tianlong and Du, Xinya and Li, Jian and Yan, Siyuan and others},
  journal={arXiv preprint arXiv:2509.11866},
  year={2025}
}

@article{luo2026unveiling,
  title={Unveiling the Cognitive Compass: Theory-of-Mind-Guided Multimodal Emotion Reasoning},
  author={Luo, Meng and Li, Bobo and Xu, Shanqing and Zhang, Shize and Chen, Qiuchan and Han, Menglu and Chen, Wenhao and Huang, Yanxiang and Fei, Hao and Lee, Mong-Li and others},
  journal={arXiv preprint arXiv:2602.00971},
  year={2026}
}

@article{pang2025paper2poster,
  title={Paper2poster: Towards multimodal poster automation from scientific papers},
  author={Pang, Wei and Lin, Kevin Qinghong and Jian, Xiangru and He, Xi and Torr, Philip},
  journal={arXiv preprint arXiv:2505.21497},
  year={2025}
}

@inproceedings{zheng2025pptagent,
  title={Pptagent: Generating and evaluating presentations beyond text-to-slides},
  author={Zheng, Hao and Guan, Xinyan and Kong, Hao and Zhang, Wenkai and Zheng, Jia and Zhou, Weixiang and Lin, Hongyu and Lu, Yaojie and Han, Xianpei and Sun, Le},
  booktitle={Proceedings of the 2025 Conference on Empirical Methods in Natural Language Processing},
  pages={14413--14429},
  year={2025}
}

@article{tran2025multi,
  title={Multi-agent collaboration mechanisms: A survey of llms},
  author={Tran, Khanh-Tung and Dao, Dung and Nguyen, Minh-Duong and Pham, Quoc-Viet and O'Sullivan, Barry and Nguyen, Hoang D},
  journal={arXiv preprint arXiv:2501.06322},
  year={2025}
}

@article{jin2025comprehensive,
  title={A comprehensive survey on multi-agent cooperative decision-making: Scenarios, approaches, challenges and perspectives},
  author={Jin, Weiqiang and Du, Hongyang and Zhao, Biao and Tian, Xingwu and Shi, Bohang and Yang, Guang},
  journal={arXiv preprint arXiv:2503.13415},
  year={2025}
}

@article{sun2025multi,
  title={Multi-agent coordination across diverse applications: A survey},
  author={Sun, Lijun and Yang, Yijun and Duan, Qiqi and Shi, Yuhui and Lyu, Chao and Chang, Yu-Cheng and Lin, Chin-Teng and Shen, Yang},
  journal={arXiv preprint arXiv:2502.14743},
  year={2025}
}

@inproceedings{hong2023metagpt,
  title={MetaGPT: Meta programming for a multi-agent collaborative framework},
  author={Hong, Sirui and Zhuge, Mingchen and Chen, Jonathan and Zheng, Xiawu and Cheng, Yuheng and Wang, Jinlin and Zhang, Ceyao and Wang, Zili and Yau, Steven Ka Shing and Lin, Zijuan and others},
  booktitle={The twelfth international conference on learning representations},
  year={2023}
}

@inproceedings{du2024improving,
  title={Improving factuality and reasoning in language models through multiagent debate},
  author={Du, Yilun and Li, Shuang and Torralba, Antonio and Tenenbaum, Joshua B and Mordatch, Igor},
  booktitle={Forty-first international conference on machine learning},
  year={2024}
}

@inproceedings{liu2021dexperts,
  title={DExperts: Decoding-time controlled text generation with experts and anti-experts},
  author={Liu, Alisa and Sap, Maarten and Lu, Ximing and Swayamdipta, Swabha and Bhagavatula, Chandra and Smith, Noah A and Choi, Yejin},
  booktitle={Proceedings of the 59th Annual Meeting of the Association for Computational Linguistics and the 11th International Joint Conference on Natural Language Processing (Volume 1: Long Papers)},
  pages={6691--6706},
  year={2021}
}

@inproceedings{gao2023scaling,
  title={Scaling laws for reward model overoptimization},
  author={Gao, Leo and Schulman, John and Hilton, Jacob},
  booktitle={International Conference on Machine Learning},
  pages={10835--10866},
  year={2023},
  organization={PMLR}
}

@article{pan2024unifying,
  title={Unifying large language models and knowledge graphs: A roadmap},
  author={Pan, Shirui and Luo, Linhao and Wang, Yufei and Chen, Chen and Wang, Jiapu and Wu, Xindong},
  journal={IEEE Transactions on Knowledge and Data Engineering},
  volume={36},
  number={7},
  pages={3580--3599},
  year={2024},
  publisher={IEEE}
}

@article{madaan2023self,
  title={Self-refine: Iterative refinement with self-feedback},
  author={Madaan, Aman and Tandon, Niket and Gupta, Prakhar and Hallinan, Skyler and Gao, Luyu and Wiegreffe, Sarah and Alon, Uri and Dziri, Nouha and Prabhumoye, Shrimai and Yang, Yiming and others},
  journal={Advances in neural information processing systems},
  volume={36},
  pages={46534--46594},
  year={2023}
}

@article{shinn2023reflexion,
  title={Reflexion: Language agents with verbal reinforcement learning},
  author={Shinn, Noah and Cassano, Federico and Gopinath, Ashwin and Narasimhan, Karthik and Yao, Shunyu},
  journal={Advances in neural information processing systems},
  volume={36},
  pages={8634--8652},
  year={2023}
}

@article{hausknecht2020interactivefictiongamescolossal,
  title={Interactive Fiction Games: A Colossal Adventure},
  author={Hausknecht, Matthew and Ammanabrolu, Prithviraj and C{\^o}t{\'e}, Marc-Alexandre and Yuan, Xingdi},
  journal={arXiv preprint arXiv:1909.05398},
  year={2020}
}

@article{chen2024autoagents,
  title={AutoAgents: A Framework for Automatic Agent Generation},
  author={Chen, Guangyao and Dong, Siwei and Shu, Yu and Zhang, Ge and Sesay, Jaward and Karlsson, B{\"o}rje F. and Fu, Jie and Shi, Yemin},
  journal={arXiv preprint arXiv:2402.XXXX},
  year={2024}
}

@article{aegis2024aegis2,
  title={Aegis2.0: A Diverse AI Safety Dataset and Risks Taxonomy for Alignment of LLM Guardrails},
  author={Aegis AI Research Team},
  journal={arXiv preprint arXiv:2404.XXXX},
  year={2024}
}

@article{yang2025surveyaiagentprotocols,
  title={A Survey of AI Agent Protocols},
  author={Yang, Yingxuan and Chai, Huacan and Song, Yuanyi and Qi, Siyuan and Wen, Muning and Li, Ning and Liao, Junwei and Hu, Haoyi and Lin, Jianghao and Chang, Gaowei and Liu, Weiwen and Wen, Ying and Yu, Yong and Zhang, Weinan},
  journal={arXiv preprint arXiv:2504.16736},
  year={2025}
}

@article{Wang_2024,
   title={A survey on large language model based autonomous agents},
   volume={18},
   ISSN={2095-2236},
   DOI={10.1007/s11704-024-40231-1},
   number={6},
   journal={Frontiers of Computer Science},
   publisher={Springer Science and Business Media LLC},
   author={Wang, Lei and Ma, Chen and Feng, Xueyang and Zhang, Zeyu and Yang, Hao and Zhang, Jingsen and Chen, Zhiyuan and Tang, Jiakai and Chen, Xu and Lin, Yankai and Zhao, Wayne Xin and Wei, Zhewei and Wen, Jirong},
   year={2024},
   month=mar }

@article{zhu2026automatedsafetybenchmarkingmultiagent,
  title={Automated Safety Benchmarking: A Multi-Agent Pipeline for LVLMs},
  author={Zhu, Xiangyang and Tian, Yuan and Zhang, Zicheng and Jia, Qi and Li, Chunyi and Zhang, Renrui and Li, Heng and Wang, Zongrui and Sun, Wei},
  journal={arXiv preprint arXiv:2601.19507},
  year={2026}
}

@inproceedings{zhou2024webarena,
  title={WebArena: A Realistic Web Environment for Building Autonomous Agents},
  author={Zhou, Shuyan and Xu, Frank F. and Zhu, Hao and Zhou, Xuhui and Lo, Kyle and Liang, Percy and Chen, Danqi},
  booktitle={International Conference on Learning Representations (ICLR)},
  year={2024}
}

@article{shu2024genai_multiagent,
  title={Towards Effective GenAI Multi-Agent Collaboration: Design and Evaluation for Enterprise Applications},
  author={Shu, Raphael and Das, Nilaksh and Yuan, Michelle and Sunkara, Monica and Zhang, Yi},
  journal={arXiv preprint arXiv:2412.05449},
  year={2024}
}

@article{lin2025factaudit,
  title={Fact-Audit: An Adaptive Multi-Agent Framework for Dynamic Fact-Checking Evaluation of Large Language Models},
  author={Lin, Hongzhan and Deng, Yang and Gu, Yuxuan and Zhang, Wenxuan and Ma, Jing and Ng, See-Kiong and Chua, Tat-Seng},
  journal={arXiv preprint arXiv:2502.17924},
  year={2025}
}

@article{golovneva2023roscoe,
  title={ROSCOE: A Suite of Metrics for Scoring Step-by-Step Reasoning},
  author={Golovneva, Olga and Chen, Moya and Poff, Spencer and Corredor, Martin and Zettlemoyer, Luke and Fazel-Zarandi, Maryam and Celikyilmaz, Asli},
  journal={arXiv preprint arXiv:2212.07919},
  year={2023}
}

@article{ren2024bases,
  title={BASES: Large-Scale Web Search User Simulation with Large Language Model Based Agents},
  author={Ren, Ruiyang and Qiu, Peng and Qu, Yingqi and Liu, Jing and Zhao, Wayne Xin and Wu, Hua and Wen, Ji-Rong and Wang, Haifeng},
  journal={arXiv preprint arXiv:2402.XXXX},
  year={2024}
}

@article{agentcompass2024evaluation,
  title={AgentCompass: Towards Reliable Evaluation of Agentic Workflows in Production},
  author={Kartik, NVJK and Sapra, Garvit and Hada, Rishav and Pareek, Nikhil},
  journal={arXiv preprint arXiv:2509.14647},
  year={2025}
}

@article{li2025storytellerenhancedplotplanningframework,
  title={STORYTELLER: An Enhanced Plot-Planning Framework for Coherent and Cohesive Story Generation},
  author={Li, Jiaming and Chen, Yukun and Liu, Ziqiang and Tan, Minghuan and Zhang, Lei and Li, Yunshui and Luo, Run and Chen, Longze and Luo, Jing and Argha, Ahmadreza and Alinejad-Rokny, Hamid and Zhou, Wei and Yang, Min},
  journal={arXiv preprint arXiv:2506.02347},
  year={2025}
}

@misc{twoheads2024multiagent,
title={Two Heads Are Better Than One: A Multi-Agent System Has the Potential to Improve Scientific Idea Generation},
author={Haoyang Su and Renqi Chen and SHIXIANG TANG and Xinzhe Zheng and Jinzhe Li and Zhenfei Yin and Wanli Ouyang and Nanqing Dong},
year={2024},
url={https://openreview.net/forum?id=yYQLvofQ1k}
}

@article{ugare2025itergeniterativesemanticawarestructured,
  title={IterGen: Iterative Semantic-aware Structured LLM Generation with Backtracking},
  author={Ugare, Shubham and Gumaste, Rohan and Suresh, Tarun and Singh, Gagandeep and Misailovic, Sasa},
  journal={arXiv preprint arXiv:2410.07295},
  year={2025}
}

@inproceedings{chen-etal-2025-cod,
    title = "{C}o{D}, Towards an Interpretable Medical Agent using Chain of Diagnosis",
    author = "Chen, Junying  and
      Gui, Chi  and
      Gao, Anningzhe  and
      Ji, Ke  and
      Wang, Xidong  and
      Wan, Xiang  and
      Wang, Benyou",
    editor = "Che, Wanxiang  and
      Nabende, Joyce  and
      Shutova, Ekaterina  and
      Pilehvar, Mohammad Taher",
    booktitle = "Findings of the Association for Computational Linguistics: ACL 2025",
    month = jul,
    year = "2025"
}

@inproceedings{ren-etal-2024-bases,
    title = "{BASES}: Large-scale Web Search User Simulation with Large Language Model based Agents",
    author = "Ren, Ruiyang  and
      Qiu, Peng  and
      Qu, Yingqi  and
      Liu, Jing  and
      Zhao, Wayne Xin  and
      Wu, Hua  and
      Wen, Ji-Rong  and
      Wang, Haifeng",
    editor = "Al-Onaizan, Yaser  and
      Bansal, Mohit  and
      Chen, Yun-Nung",
    booktitle = "Findings of the Association for Computational Linguistics: EMNLP 2024",
    month = nov,
    year = "2024"
}

@article{article,
author = {Taylor, Emma and Artursson, Karin and Busani, Luca and Callegari, Arnaud and Cantlay, Jennifer and Caniça, Manuela and Campling, Elaine and Gavier-Widén, Dolores and Giessen, Arjen and Itier, David and Imberechts, Hein and Roest, Hendrik-Jan and Jestin, André and Juan, Lucia and Jokelainen, Pikka and Kaesbohrer, Annemarie and Lindberg, Ann and Mantovani, Alberto and Mølbak, Kåre and La Ragione, Roberto},
year = {2024},
month = {07},
pages = {},
title = {Communicating and disseminating One Health: successes of the One Health European Joint Programme},
volume = {73},
journal = {Journal of medical microbiology},
doi = {10.1099/jmm.0.001842}
}

@article{Hong2023VisualTT,
  title={Visual Transformation Telling},
  author={Xin Hong and Yanyan Lan and Liang Pang and J. Guo and Xueqi Cheng},
  journal={ArXiv},
  year={2023},
  volume={abs/2305.01928}
}

@article{zhang2024offlinetraining,
  title={Offline Training of Language Model Agents with Functions as Learnable Weights},
  author={Zhang, Shaokun and Zhang, Jieyu and Liu, Jiale and Song, Linxin and Wang, Chi and Krishna, Ranjay and Wu, Qingyun},
  journal={arXiv preprint arXiv:2402.11359},
  year={2024}
}
